\documentclass[runningheads]{llncs}

\usepackage{eccv}

\usepackage{eccvabbrv}

\usepackage{graphicx}
\usepackage{booktabs}

\usepackage[accsupp]{axessibility}  

\usepackage{hyperref}

\usepackage{orcidlink}
\usepackage{array}        
\usepackage{makecell}     
\usepackage{multirow}     
\usepackage{multicol}     
\usepackage{colortbl}     
\usepackage[table]{xcolor}
\usepackage{siunitx}      
\usepackage{listings}
\usepackage{xcolor}
\usepackage{cuted}
\usepackage{graphicx}
\usepackage{caption}
\usepackage{bm}
\usepackage{amssymb}
\usepackage{pifont}

\begin{document}

\title{Refinement via Regeneration: \\ Enlarging Modification Space Boosts Image Refinement in Unified Multimodal Models}

\titlerunning{Refinement via Regeneration}

\author{Jiayi Guo\inst{1,2} \and
Linqing Wang\inst{2} \and Jiangshan Wang\inst{1,2} \and Yang Yue\inst{1} \and Zeyu Liu\inst{1} \and Zhiyuan Zhao\inst{2} \and Qinglin Lu\inst{2\dagger} \and Gao Huang\inst{1\dagger} \and Chunyu Wang\inst{2\dagger}\\ \vspace{2mm}{\small
    $^{1}$Tsinghua University\ \ \
    $^{2}$Tencent HY}\\
    {\small$^{\ddagger}$Corresponding Authors}\\
     {\small \textbf{\color{magenta}\url{https://github.com/LeapLabTHU/RvR}}}
}

\authorrunning{J. Guo et al.}
\institute{}


\maketitle


\begin{center}
    \centering
  \includegraphics[width=\linewidth]{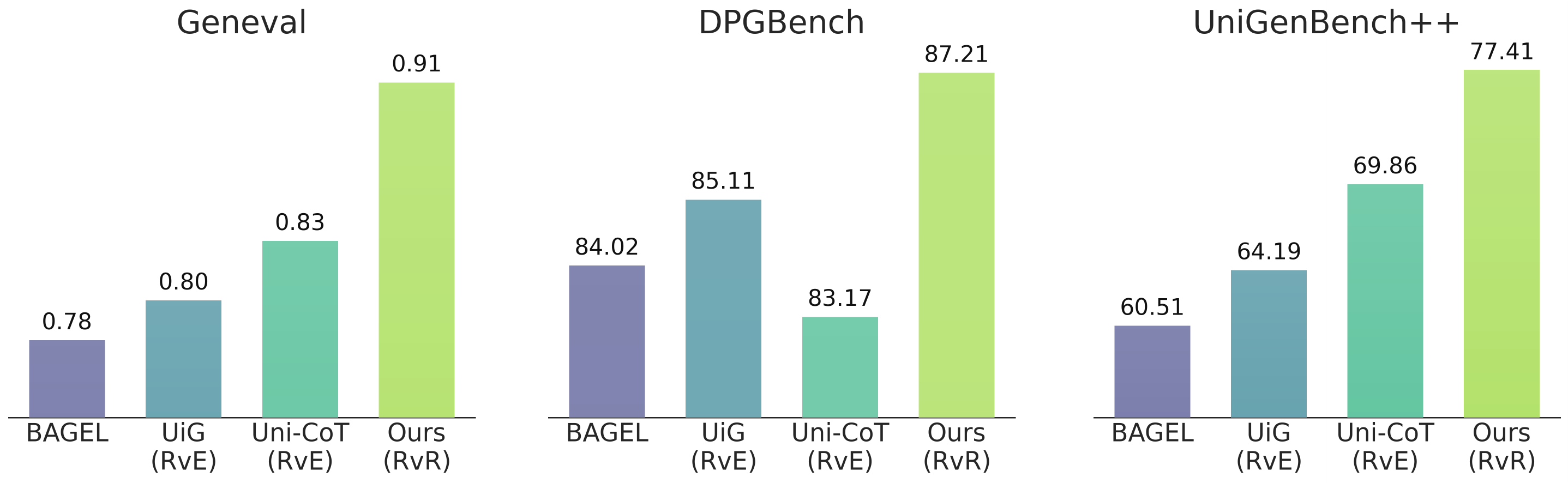}\vspace{-2mm}
  \captionof{figure}{
\textbf{Refinement via Regeneration (RvR) largely improves text-to-image generation.}
Compared with the base unified multimodal model (UMM) BAGEL~\cite{bagel} and existing refinement-via-editing (RvE) methods, RvR achieves consistently better performance across Geneval~\cite{geneval}, DPGBench~\cite{ella}, and UniGenBench++~\cite{unigenbench++}.
 }\vspace{-2mm}
  \label{fig:teaser}
\end{center}

\begin{abstract}

Unified multimodal models (UMMs) integrate visual understanding and generation within a single framework. For text-to-image (T2I) tasks, this unified capability allows UMMs to refine outputs after their initial generation, potentially extending the performance upper bound. Current UMM-based refinement methods primarily follow a refinement-via-editing (RvE) paradigm, where UMMs produce editing instructions to modify misaligned regions while preserving aligned content. However, editing instructions often describe prompt–image misalignment only coarsely, leading to incomplete refinement. Moreover, pixel-level preservation, though necessary for editing, unnecessarily restricts the effective modification space for refinement. To address these limitations, we propose \textbf{R}efinement \textbf{v}ia \textbf{R}egeneration (\textbf{RvR}), a novel framework that reformulates refinement as conditional image regeneration rather than editing. Instead of relying on editing instructions and enforcing strict content preservation, RvR regenerates images conditioned on the target prompt and the semantic tokens of the initial image, enabling more complete semantic alignment with a larger modification space. Extensive experiments demonstrate the effectiveness of RvR, improving Geneval from 0.78 to 0.91, DPGBench from 84.02 to 87.21, and UniGenBench++ from 61.53 to 77.41. 



\end{abstract}
 
\section{Introduction}
\label{sec:intro}

\begin{figure}[t]
    \centering
    \includegraphics[width=\linewidth]{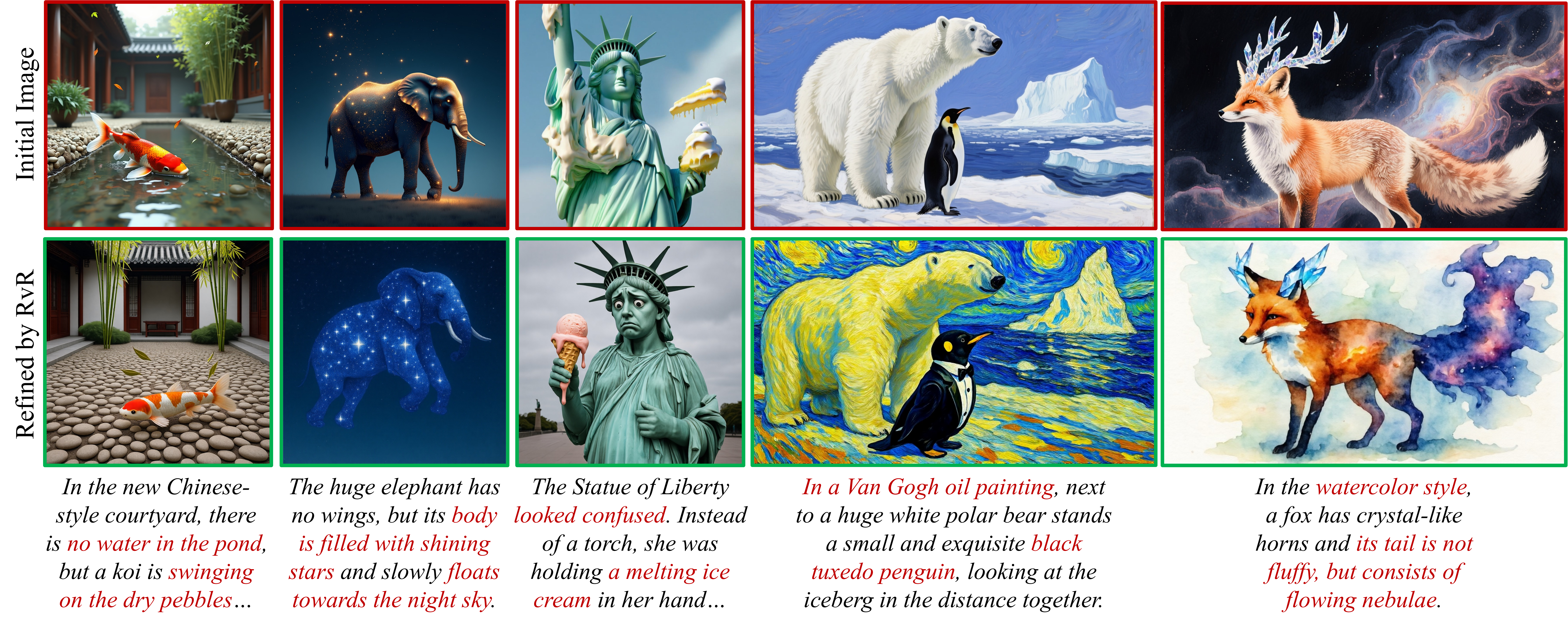}  \vspace{-6mm}
    \caption{\textbf{Qualitative examples before and after RvR refinement.}
    }\vspace{-4mm}
    \label{fig:illu2}
\end{figure}

\begin{figure}[t]
    \centering
    \includegraphics[width=\linewidth]{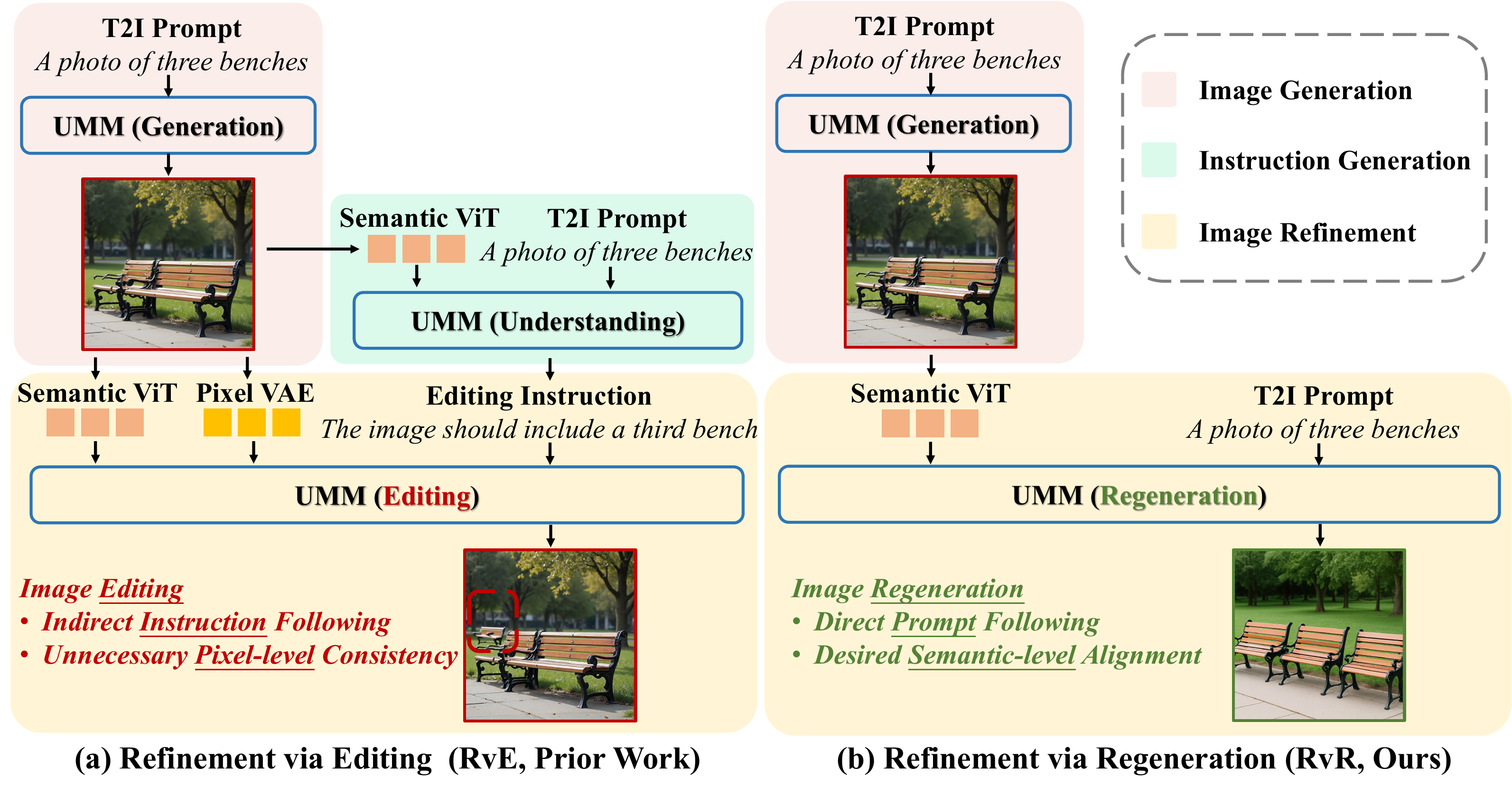}  \vspace{-6mm}
    \caption{\textbf{Comparison between (a) prior Refinement via Editing (RvE) and (b) our Refinement via Regeneration (RvR).} RvE requires precise instruction generation and content consistency for unedited regions, while RvR discards these unnecessary constraints, enlarging modification space for better refinement. 
    }\vspace{-6mm}
    \label{fig:illu1}
\end{figure}

Modern text-to-image (T2I) generation models~\cite{gan,ddpm,sd,dit,bagel,qwenimage,hunyuanimage3} have made remarkable progress in synthesizing high-fidelity images from natural language. Nevertheless, reliably following complex prompts remains a major challenge, particularly when prompts involve multiple objects, diverse attributes, or fine-grained relationships~\cite{geneval,ella,unigenbench++}.
To improve prompt--image alignment, recent studies have explored
refinement approaches based on unified multimodal models (UMMs)~\cite{bagel,janus,januspro,janusflow,emu,blip3o,blip3onext,metaqueries}. By integrating image understanding, generation, and editing within a single framework, UMMs can analyze a generated image in the context of a target prompt and iteratively improve it.
Most existing UMM-based refinement methods instantiate this idea through \emph{Refinement via Editing} (RvE)~\cite{unicot,uig,irg}.
As illustrated in~\cref{fig:illu1}(a), given a generated image and a target prompt, a UMM first produces an editing instruction that summarizes their semantic mismatch through image understanding, and then performs image editing conditioned on this instruction to refine the image.

Despite the feasibility of editing-based refinement, we argue that RvE suffers from several inherent
limitations that bound its performance ceiling.
First, the intermediate editing instruction is often a coarse and incomplete description of the semantic gap between the current image and the target prompt~\cite{magicbrush,ultraedit,anyedit}.
As shown in~\cref{fig:illu1}(a), an instruction such as ``include a third bench'' addresses only part of the mismatch while inevitably ignoring several other necessary corrections, \eg, removing extra armrests, adjusting the layout, or harmonizing the appearance of existing benches.
Consequently, the subsequent editing step is guided by an incomplete specification and may yield suboptimal refinement.
Second, the editing formulation enforces strict pixel-level consistency in unedited regions by design.
While this constraint is essential for image editing, it is unnecessary for image refinement and directly restricts the effective modification space.
For example, in~\cref{fig:illu1}(a), preserving the original content leaves insufficient room to insert an additional bench, resulting in an unnaturally small and visually low-quality insertion.
In contrast, refinement should prioritize semantic correctness and overall visual plausibility of the final image---in this example, producing three natural, well-composed benches---even if achieving this requires broader structural changes beyond localized edits.

Motivated by these observations, we propose \emph{Refinement via Regeneration} (RvR), a novel framework that reformulates image refinement from a regeneration perspective, as illustrated in~\cref{fig:illu1}(b). Rather than relying on intermediate editing instructions or enforcing pixel-level consistency with the input image, RvR conditions directly on the target prompt together with the semantic tokens of the initial image, treating refinement as another round of image generation. Removing editing-specific constraints substantially enlarges the effective modification space, allowing the model to revise any regions that hinder prompt satisfaction. As shown in~\cref{fig:illu1}(b), RvR produces refined images that are semantically correct, spatially coherent, and better aligned with the target prompt.

This regeneration-based refinement mechanism is supported by a training scheme that enlarges the effective modification space from both data and pipeline perspectives. At the data level, we construct supervision from independently generated T2I samples with varying degrees of prompt alignment, rather than from editing pairs that enforce strict content consistency. Consequently, the aligned image need not be an edited version of the misaligned image; the supervision emphasizes semantic correction toward the target prompt without imposing pixel-level correspondence.
At the pipeline level, RvR further simplifies the refinement pipeline by discarding pixel-level VAE conditioning and relying only on semantic ViT representations of the input image. This design allows the model to revise image content guided by high-level semantics, instead of being biased toward appearance preservation.

Extensive experiments across multiple benchmarks demonstrate that RvR consistently delivers substantial gains in T2I generation, with particularly strong improvements in prompt--image alignment for semantically complex prompts.
Specifically, RvR boosts Geneval from 0.78 to 0.91, DPGBench from 84.02 to 87.21, and UniGenBench++ from 61.53 to 77.41. 
These results indicate that regeneration, rather than editing, provides a more effective and principled foundation for image refinement in unified multimodal models.

\section{Related Work}
\label{sec:related}

\noindent{\textbf{Unified multimodal models.}}
Early T2I models~\cite{dall-e,dalle2,sd,flux,glide,faceclip,pfd,pixartalpha,smoothdiffusion,rfsolver}, such as Stable Diffusion~\cite{sd,sdxl,sd3} and FLUX~\cite{flux}, typically rely on frozen text encoders (e.g., CLIP~\cite{clip} or T5~\cite{t5}) to map prompts into static embeddings, which fundamentally limits their ability to perform rich semantic understanding. To overcome this bottleneck, recent studies have proposed unified multimodal models (UMMs)~\cite{bagel,janus,januspro,janusflow,blip3o,blip3onext,metaqueries} that tightly integrate large language models~\cite{llama3,gpt4} or vision-language models~\cite{llava,qwenvl} with visual generation within a single framework, substantially enhancing prompt comprehension and instruction following~\cite{showo, chameleon, bagel, janus, blip3o, promptenhancer}. Existing UMMs explore diverse design choices along architecture, representation, and training objectives, including early-fusion or single-Transformer architectures~\cite{showo, chameleon}, unified image tokenization and multi-granularity modeling~\cite{tokenflow, seedx}, as well as the unification of autoregressive language modeling with diffusion- or flow-based image synthesis~\cite{transfusion, janusflow}. Moreover, large-scale unified pretraining has been shown to induce strong emergent multimodal capabilities~\cite{bagel, januspro, blip3o}, while recent efforts further extend UMMs toward self-enhancement and long-context video--language modeling~\cite{illume, lwm}.

\smallskip
\noindent{\textbf{Refinement for T2I generation.}}
Refinement~\cite{sld,img,uig,unicot,irg} aims to improve a preliminary T2I sample by reducing prompt--image mismatches, especially for compositional prompts~\cite{geneval,ella,unigenbench++}. A first line of work augments \emph{conventional} diffusion models—whose prompt encoders lack strong reasoning—by coupling them with external LLM/VLM understanding in a closed loop: SLD detects objects, analyzes inconsistencies, and applies training-free, object-level latent edits (add/move/replace) guided by an LLM controller~\cite{sld,masterllm}, while IMG uses an MLLM to diagnose misalignments and calibrates diffusion conditioning via an implicit aligner to enable regeneration without explicit editing operations or extra data~\cite{img}. In contrast, more recent efforts build refinement \emph{inside} UMMs that natively support both multimodal understanding and image generation, and thus can turn refinement into a model-internal reasoning--generation loop~\cite{uig,unicot,irg}. These UMM-based methods largely follow a \emph{refinement-via-editing} paradigm, where the unified model first interprets the prompt--image mismatch and then refines the image by generating explicit or implicit editing actions that are executed through its image editing or synthesis capability. For example, UiG leverages image editing as an explicit interface to inject the unified model's understanding into step-wise visual modifications~\cite{uig}. Other approaches interleave multimodal reasoning and image updates more loosely, such as alternating textual reflection with image synthesis to gradually correct errors and enhance visual details~\cite{irg}, or organizing refinement as a unified chain-of-thought across text and vision to maintain coherent visual-state transitions throughout the editing process~\cite{unicot}. 
\section{Refinement via Regeneration}
\label{sec:rvr}

In this section, we first introduce the preliminaries of unified multimodal models (UMMs) and the existing refinement-via-editing (RvE) paradigm. We then analyze the unnecessary constraints imposed by RvE that bound the performance ceiling of prompt--image alignment.
Finally, we present our \emph{Refinement via Regeneration} (RvR) pipeline, which removes these editing-specific constraints. This design enlarges the feasible modification space and enables the model to produce refined images that better align with the target prompt.

\subsection{Background}\label{sec:bg}
\textbf{Unified multimodal models} (UMMs) integrate image understanding, image generation, and image editing within a single generative framework. Representative UMMs, such as BAGEL~\cite{bagel}, typically combine specialized visual encoders with expert modules for both understanding and generation.

For image understanding, a semantic visual encoder, usually parameterized by a pre-trained Vision Transformer~\cite{vit} (ViT), extracts high-level semantic features $Z_{\rm ViT}$ from an input image. These semantic tokens are fed into the multimodal backbone for joint reasoning with text. For image synthesis, a variational autoencoder~\cite{vae} (VAE) maps the image into low-level latent tokens $Z_{\rm VAE}$, which supports generative modeling through flow matching~\cite{rf}. Equipped with both semantic visual representations and generative image latents, the UMM backbone $\mathcal{M}$ supports the following tasks:
\begin{itemize}
  \item \textbf{Image understanding.}
  Given an input image $I$ and a text query $T$, \eg, a question about the input image~\cite{vqa},
  the model $\mathcal{M}$ generates a textual response $\hat{T}$:
  \begin{equation}
  \hat{T} = \mathcal{M}\bigl(T,\, Z_{\rm ViT}\bigr),\label{eq:und}
  \end{equation}
  where $Z_{\rm ViT}$ denotes the semantic visual tokens extracted from $I$.

  \item \textbf{Text-to-image generation.}
  Given a text prompt $T_{\rm{prompt}}$, the model $\mathcal{M}$ synthesizes an image $\hat{I}$ that follows the prompt:
  \begin{equation}
  \hat{I} = \mathcal{M}(T_{\rm prompt}).
  \end{equation}

  \item \textbf{Image editing.}
  Given an input image $I$ and an editing instruction $T_{\rm{edit}}$,
  the model produces an edited image $\hat{I}'$ by modifying $I$ according to $T_{\rm{edit}}$:
  \begin{equation}
  \hat{I}' =
  \mathcal{M}\bigl(T_{\rm{edit}},\, Z_{\rm ViT},\, Z_{\rm VAE}\bigr),\label{eq:edit}
  \end{equation}
  where $Z_{\rm VAE}$ denotes the low-level VAE tokens of $I$.
\end{itemize}

\noindent\textbf{Refinement via Editing} (RvE)~\cite{uig,unicot,irg}. 
By jointly supporting image understanding and image generation, UMMs are well-suited for image refinement tasks: analyzing the semantic misalignment between an image and its target prompt, and refining the image to better follow the prompt. 
Existing methods typically adopt a refinement-via-editing (RvE) paradigm, which decomposes refinement into two stages: instruction generation and image editing.

In the first stage, the model $\mathcal{M}$ compares an input image $I$ with the target prompt $T_{\rm prompt}$ and generates an editing instruction $\hat{T}_{\rm edit}$ that describes how the image should be modified to better satisfy the prompt. 
This step follows the image understanding formulation in Eq.~\ref{eq:und}:
\begin{equation}
  \hat{T}_{\rm edit}
  = \mathcal{M}\bigl(T_{\rm prompt},\, Z_{\rm ViT}\bigr),\label{eq:rve1}
\end{equation}
where $Z_{\rm ViT}$ denotes the semantic visual tokens of $I$.

In the second stage, the model edits the image $I$ according to the generated instruction $\hat{T}_{\rm edit}$ to produce a refined image $\hat{I}'$:
\begin{equation}
  \hat{I}'
  = \mathcal{M}\bigl(\hat{T}_{\rm edit},\, Z_{\rm ViT},\, Z_{\rm VAE}\bigr).\label{eq:rve2}
\end{equation}
This follows the standard image editing formulation in Eq.~\ref{eq:edit}, except that the instruction is produced by the model itself rather than provided externally.

During training, the instruction generation stage is supervised with the ground-truth editing instruction $T_{\rm edit}$ and optimized using an autoregressive text loss:
\begin{equation}
\mathcal{L}_{\rm text}
= \mathbb{E}
\Big[
-\log p_{\mathcal M}(T_{\rm edit}\mid T_{\rm prompt}, Z_{\rm ViT})
\Big],
\end{equation}
where $p_{\mathcal M}$ denotes the text output distribution defined by $\mathcal M$.

The image editing stage is trained on triplets $\langle I, I', T_{\rm edit}\rangle$, where $I$ is the original image, $I'$ is the target edited image, and $T_{\rm edit}$ is the corresponding editing instruction. 
Image synthesis is formulated under rectified flow (RF)~\cite{rf}. Let $\bm{x}_0 = Z'_{\rm VAE}$ denote the VAE tokens of the target image $I'$. RF defines a linear interpolation between the clean target $\bm{x}_0$ and a Gaussian noise $\bm{x}_1 \sim \mathcal{N}(\bm{0}, \bm{I})$:
\begin{equation}
\bm{x}_t = (1-t)\bm{x}_0 + t \bm{x}_1,
\quad t \sim \mathcal{U}(0,1).
\label{eq:rf_interp}
\end{equation}
Given the noisy tokens $\bm{x}_t$ and the conditioning context, including the editing instruction $T_{\rm edit}$ together with the ViT and VAE tokens of the original image $I$, the model predicts the velocity field $v_{\theta}(\bm{x}_t, \cdot)$ and is trained with the flow matching (FM) objective:
\begin{equation}
\mathcal{L}_{\mathrm{FM}}
= \mathbb{E}
\Big[
\big\|
v_{\theta}\bigl(\bm{x}_t, T_{\rm edit}, Z_{\rm ViT}, Z_{\rm VAE}\bigr)
- (\bm{x}_1 - \bm{x}_0)
\big\|_2^2
\Big].
\label{eq:rf_loss}
\end{equation}

\begin{figure}[t]
    \centering
    \includegraphics[width=0.95\linewidth]{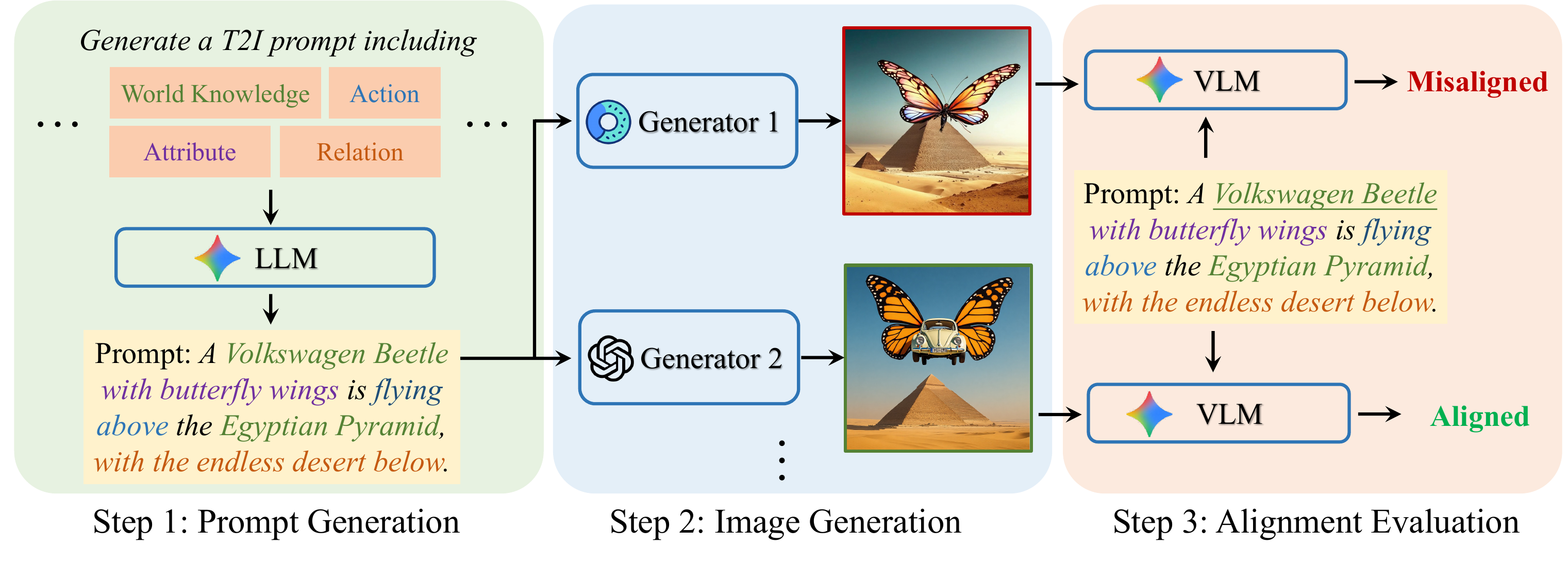} \vspace{-3mm}
    \caption{\textbf{Data construction pipeline for RvR.} Step 1: An LLM generates prompts based on randomly selected semantic dimensions. Step 2: Multiple T2I generators independently generate images. Step 3: A VLM evaluates prompt–image alignment and labels generated images as aligned or misaligned. Each final training sample is constructed as a triplet of $\langle$misaligned image, aligned image, prompt$\rangle$.
    }\vspace{-4mm}
    \label{fig:data}
\end{figure}

\begin{figure}[t]
    \centering
    \includegraphics[width=0.95\linewidth]{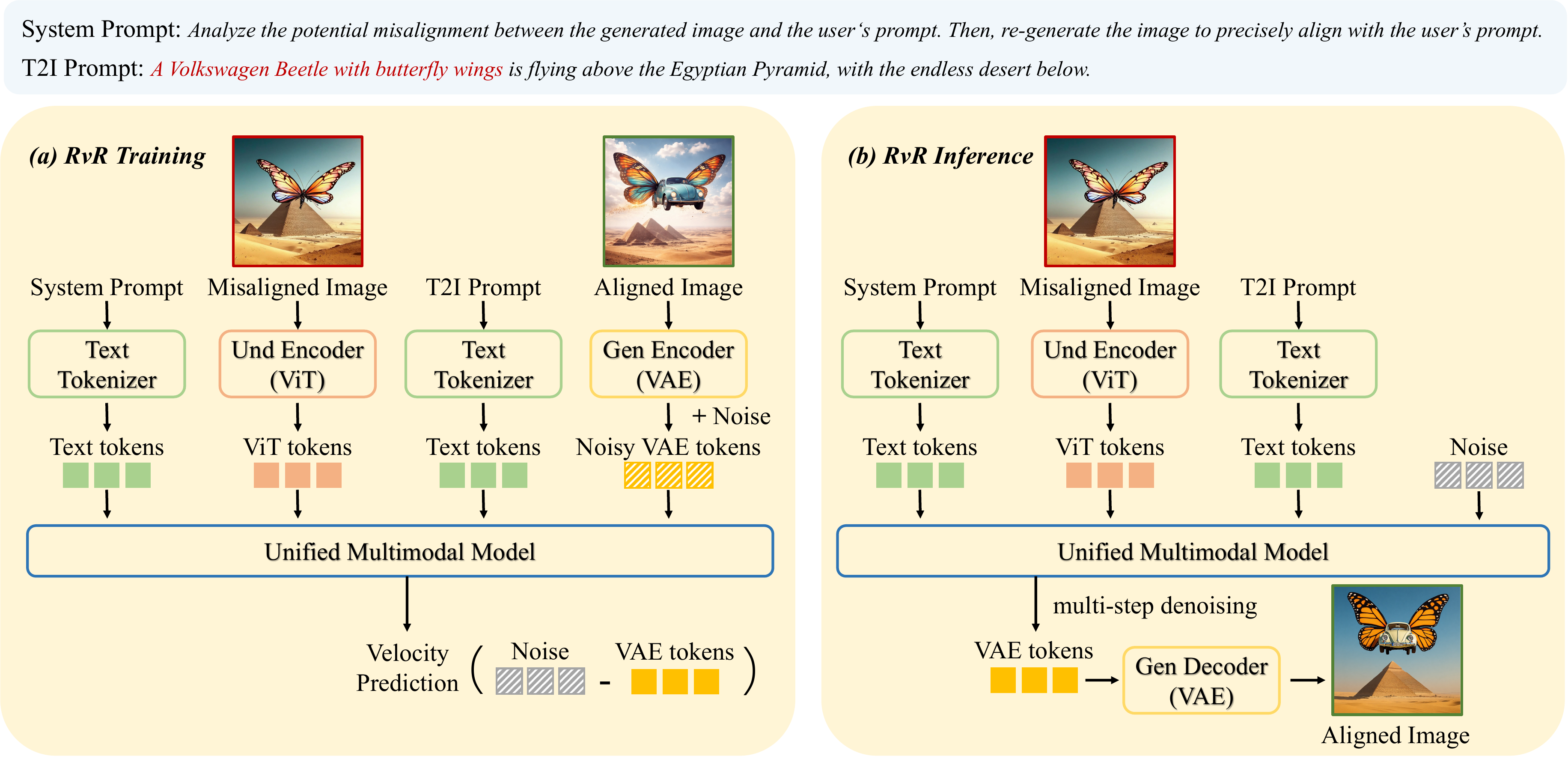}\vspace{-2mm}
    \caption{\textbf{Overview of RvR.} During training (a), the unified multimodal model (UMM) takes text tokens, ViT tokens from a misaligned image, and noisy VAE tokens from an aligned image, and learns velocity prediction for denoising. During inference (b), conditioned on the system prompt and misaligned image, the UMM denoises the noise to refined VAE tokens, which are decoded into the final aligned image. 
    }\vspace{-4mm}
    \label{fig:overview}
\end{figure}

\subsection{From RvE to RvR: Eliminating Unnecessary Constraints}

Although RvE provides a natural way to enable self-refinement with UMMs, it inherits several constraints from image editing that are unnecessary for the refinement problem.
In particular, RvE typically requires two-stage training on triplets $\langle$original image, edited image, editing instruction$\rangle$, which leads to a highly demanding data construction pipeline. The editing instruction must accurately and comprehensively describe the semantic differences between the original and edited images, while the edited image is expected to modify only the intended regions and preserve all other content. Such supervision is well aligned with the goal of image editing, where faithful local modification and content preservation are essential, but it becomes unnecessarily restrictive for image refinement.

First, the reliance on an intermediate editing instruction introduces an additional source of error. If the instruction is incomplete, ambiguous, or only partially captures the semantic gap, the subsequent editing stage is inherently limited by this imperfect intermediate representation, leading to \emph{error accumulation} across the two stages. 

More importantly, RvE constrains refinement to operate within the modification space of image editing. Because the refined image is expected to remain strongly correspondent and spatially aligned with the original image, the model is biased toward conservative, localized modifications. While such constraints are desirable for editing, they are not necessary for refinement. The goal of image refinement is simply to produce an image that better aligns with the target prompt. Therefore, formulating refinement as editing unnecessarily \emph{restricts the modification space} and limits the achievable refinement performance. To address this limitation, we propose RvR, which removes editing-specific constraints and reformulates refinement as conditional image regeneration with solely the target prompt and semantic tokens of the initial image.

\subsection{RvR Data Construction}
\label{sec:dc} 

RvR adopts a substantially simpler and more scalable data construction pipeline while maintaining high-quality supervision tightly aligned with the objective of text-to-image generation.
Specifically, RvR removes reliance on editing instructions and discards unnecessary content consistency constraints between input and output images.
As shown in~\cref{fig:data}, our data construction pipeline consists of three steps: prompt generation, image generation, and alignment evaluation.

\smallskip
\noindent\textbf{Prompt generation.} We construct a diverse prompt set that covers a wide range of semantic dimensions following~\cite{unigenbench++}.
Specifically, for each prompt we randomly select 1--5 semantic dimensions (\eg, style, world knowledge, and quantity),
and then employ a large language model (LLM, \eg, Gemini) to generate a textual prompt that simultaneously incorporates all selected dimensions.

\smallskip
\noindent\textbf{Image generation.} For each prompt, we use multiple generators (\eg, Bagel~\cite{bagel} and GPT-4o~\cite{gpt4o}) to independently synthesize candidate images.
This construction explicitly avoids content-consistency constraints, thereby \emph{enlarging both the learning and modification space} of RvR: the model is encouraged to move beyond conservative, edit-like updates and instead learn how to semantically transform a misaligned image into a more prompt-faithful one.

\smallskip
\noindent\textbf{Alignment evaluation.} 
For each candidate prompt--image pair, we use a vision--language model (VLM, \eg, Gemini) to evaluate their semantic alignment. Finally, for each prompt, we select one misaligned image and one aligned image to form a training triplet $\langle I, I', T\rangle$, where $I$ denotes the misaligned image, $I'$ the aligned image, and $T$ the prompt.

\subsection{RvR Pipeline: Training and Inference}
\label{sec:pipe}

\cref{fig:overview} illustrates the training and inference pipeline of RvR.
To guide regeneration, we design a system prompt $T_{\rm system}$:
``\emph{Analyze the potential misalignment between the generated image and the user's prompt.
Then, re-generate the image to precisely align with the user's prompt.}''
This system prompt explicitly defines the role of the UMM in image refinement: instead of editing the input image according to an intermediate instruction, the model directly regenerates a new image that better satisfies the user prompt. In this way, RvR reformulates refinement as a regeneration problem, rather than a constrained editing task.

\smallskip
\noindent\textbf{Training.}
During training, RvR takes four inputs:
the system prompt $T_{\rm system}$,
the semantic tokens $Z_{\rm ViT}$ of the misaligned image,
the user prompt $T_{\rm T2I}$,
and the noisy VAE tokens $\bm{x}_t$ of the aligned image.
Conditioned on these inputs, RvR predicts the velocity field
$v_{\theta}(\bm{x}_t, \cdot)$ and is optimized with the FM objective:
\begin{equation}
\mathcal{L}_{\mathrm{FM}}
= \mathbb{E}
\Big[
\big\|
v_{\theta}\bigl(\bm{x}_t, T_{\rm prompt}, Z_{\rm ViT}\bigr)
- (\bm{x}_1 - \bm{x}_0)
\big\|_2^2
\Big],
\label{eq:rf_loss1}
\end{equation}
where $T_{\rm prompt}=T_{\rm system}\oplus T_{\rm T2I}$ denotes the concatenated prompt used for conditioning.
Different from Eq.~\ref{eq:rf_loss}, our formulation discards $Z_{\rm VAE}$ from the conditioning context.
In editing-based refinement, these VAE tokens provide low-level information that encourages the output to remain consistent with the input image.
In contrast, regeneration aims to allow larger modifications toward better prompt alignment, and therefore removes this unnecessary prior.
We empirically validate the performance gain from removing VAE tokens in~\cref{sec:abl}.

\smallskip
\noindent\textbf{Inference.}
At inference time, given a prompt $T_{\rm prompt}$ and a misaligned image $I$,
RvR regenerates an improved image $\hat{I}'$ conditioned on semantic tokens $Z_{\rm ViT}$ extracted from $I$:
\begin{equation}
\hat{I}'
=
\mathcal{M}\bigl(T_{\rm prompt},\, Z_{\rm ViT}\bigr).
\label{eq:rvr}
\end{equation}
Compared with Eq.~\ref{eq:rve1} and Eq.~\ref{eq:rve2}, this formulation offers two key advantages.
First, RvR directly conditions on the target prompt without relying on an intermediate editing instruction, thereby avoiding error accumulation caused by incomplete or inaccurate instructions.
Second, RvR operates solely on high-level semantic representations of the input image rather than enforcing pixel-level consistency with the source image.
This removes unnecessary content-preservation constraints, enlarges the effective modification space, and allows the model to make more flexible changes toward better prompt--image alignment.
\section{Experiment}
\label{sec:exp}

\subsection{Experimental Setup}\label{sec:setup}

\noindent\textbf{Baselines and models.}
Our experiments are conducted on BAGEL~\cite{bagel}, a widely adopted base UMM in the context of the image refinement task.
We compare RvR with several representative RvE-based methods, including UiG~\cite{uig}, Uni-CoT~\cite{unicot}, and IRG~\cite{irg}.
For data construction, we use BAGEL and GPT-4o~\cite{gpt4o} to generate candidate images, and employ Gemini-2.5-Pro~\cite{gemini25pro} for prompt generation and prompt--image alignment evaluation.

\smallskip
\noindent\textbf{Implementation details.}
Our RvR pipeline is trained from BAGEL~\cite{bagel} using 16 NVIDIA H800 GPUs for 15K steps with the AdamW optimizer~\cite{adamw} and a learning rate of $1\times10^{-4}$.
The exponential moving average (EMA) decay~\cite{tarvainen2017mean} is set to 0.9999.
The training objective combines cross-entropy loss and mean squared error loss with a weight ratio of 0.25:1.
The training data consists of three parts:
(a) 100k image refinement samples constructed as described in \cref{sec:dc}, used to learn the core RvR semantic correction capability;
(b) 60k text-to-image samples from BLIP-3o~\cite{blip3o}, used to preserve basic T2I generation ability; and
(c) 1k image understanding samples from the BAGEL repository, used to maintain visual reasoning ability.
During training, image refinement, text-to-image, and image understanding samples are mixed with a ratio of 2:1:1.
During inference, we use 50 sampling steps.
Classifier-free guidance (CFG)~\cite{cfg} is applied with a text guidance scale of 4 and an image guidance scale of 2.
We evaluate RvR on three widely used T2I benchmarks: Geneval~\cite{geneval}, DPGBench~\cite{ella}, and UniGenBench++~\cite{unigenbench++}, covering prompts ranging from short object compositions to dense and complex semantics.

\begin{figure}[t]
    \centering
    \includegraphics[width=\linewidth]{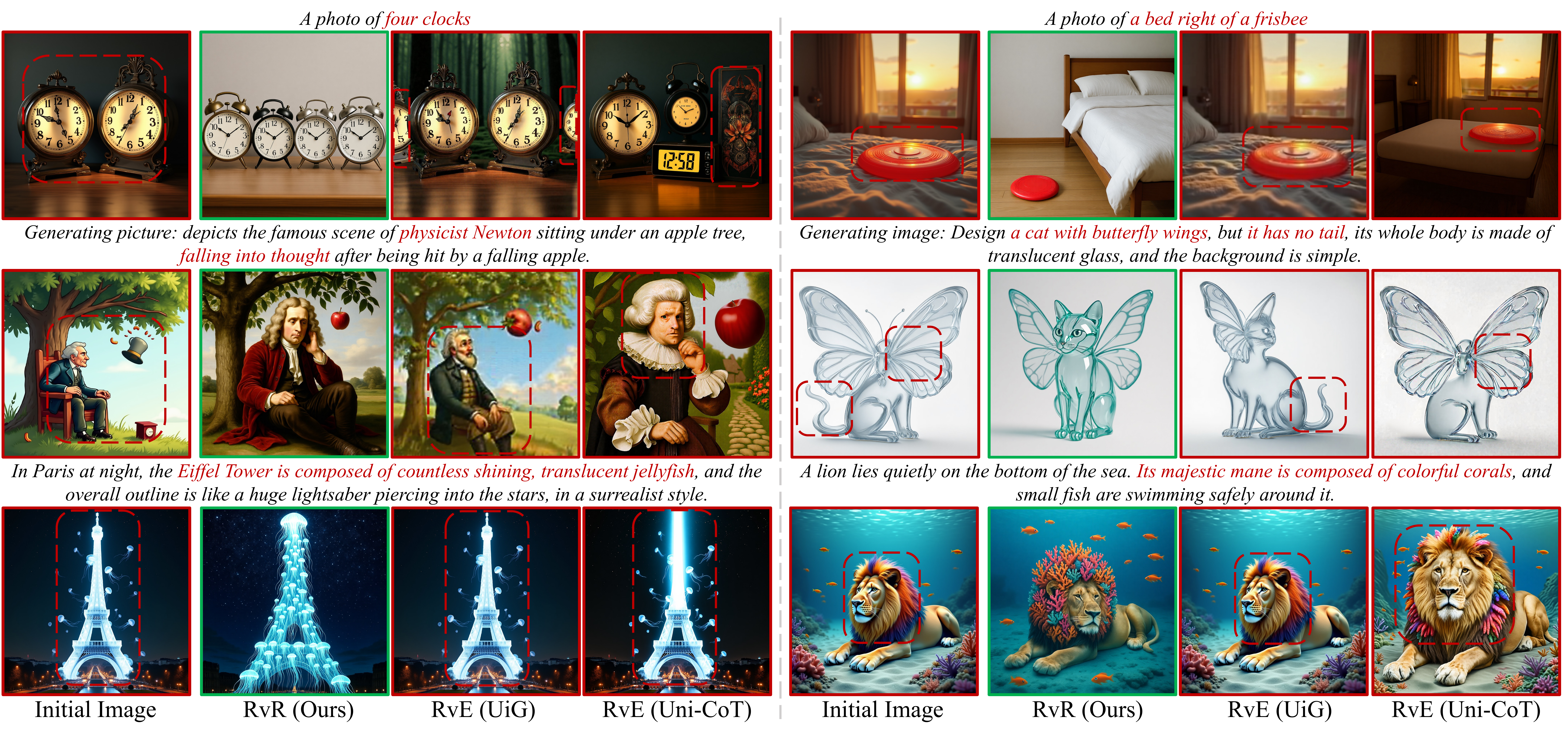}  \vspace{-6mm}
    \caption{\textbf{Qualitative comparison with RvE methods.} We compare the image refinement results of our RvR with representative RvE-based methods, UiG~\cite{uig} and Uni-CoT~\cite{unicot}. 
The results demonstrate the superiority of RvR in correcting various semantic misalignments. Red dashed boxes highlight the misaligned regions.
    }\vspace{-4mm}
    \label{fig:expr1}
\end{figure}

\subsection{Main Results}

\noindent\textbf{Qualitative comparison.}
We first present qualitative image refinement results of RvR and compare them with two representative RvE-based methods, UiG~\cite{uig} and Uni-CoT~\cite{unicot}. 
For fair comparison, the initial images for refinement are identical across methods and are synthesized by BAGEL~\cite{bagel}. 
As illustrated in~\cref{fig:expr1}, RvR shows clear advantages in correcting various semantic misalignments, including object quantity (``four clocks''), relative position (``right of''), world knowledge (``physicist Newton''), negation grammar (``has no tail''), and object composition (``composed of''). 
These improvements mainly stem from the larger modification space enabled by RvR compared with RvE-based methods.
Take the case ``A photo of a bed right of a frisbee'' as an example. 
The initial image contains an irrelevant window occupying more than half of the scene. 
During refinement, both UiG and Uni-CoT preserve this window because they are trained to maintain unedited regions. 
Such unnecessary preservation restricts the modification space required to relocate the bed and the frisbee, leading to failure cases. 
In contrast, RvR focuses solely on semantic alignment. 
It correctly identifies the key misalignment (the frisbee is placed on top of the bed) and relocates the bed to the right of the frisbee, producing a prompt-aligned result.

\smallskip
\noindent\textbf{Quantitative comparison.}
In~\cref{tab:bagel_comparison}, we evaluate RvR on three mainstream T2I benchmarks: Geneval~\cite{geneval}, DPGBench~\cite{ella}, and UniGenBench++~\cite{unigenbench++}. 
RvR achieves consistent leading performance across all benchmarks, demonstrating its effectiveness in handling both short compositional prompts and dense semantic prompts. 
Compared with the base model BAGEL~\cite{bagel}, the regeneration-based refinement significantly improves generation quality, boosting Geneval from 0.78 to 0.91, DPGBench from 84.02 to 87.21, and UniGenBench++ from 61.53 to 77.41. 
RvR also clearly outperforms editing-based refinement methods, including UiG~\cite{uig}, UniCoT~\cite{unicot}, and IRG~\cite{irg}, achieving 0.91 vs.\ 0.85 on Geneval, 87.21 vs.\ 85.11 on DPGBench, and 77.41 vs.\ 69.86 on UniGenBench++.
These gains suggest that strict pixel-level consistency constraints imposed by editing-based methods are unnecessary for effective refinement. By discarding such constraints, RvR enlarges the modification space and achieves a higher performance ceiling. 
Moreover, BAGEL-RvR also reaches state-of-the-art performance compared with existing T2I models and unified multimodal models.

\begin{table}[t]
\centering
\caption{\textbf{Quantitative comparison with T2I and refinement methods.} Our BAGEL-RvR consistently outperforms generation-only models, UMMs and RvE methods. $\dagger$ indicates our reproduced results.}\vspace{-3mm}
\label{tab:bagel_comparison}
\setlength{\tabcolsep}{2pt}
\resizebox{1\linewidth}{!}{\begin{tabular}{llccccccccc}
\toprule
\multirow{2}{*}{\textbf{Type}} & \multirow{2}{*}{\textbf{Model}} & \multicolumn{7}{c}{\textbf{Geneval$\uparrow$}} & \multirow{2}{*}{\textbf{DPGBench$\uparrow$}} & \multirow{2}{*}{\textbf{UniGenBench++$\uparrow$}}\\
&& \textbf{Single Obj.} & \textbf{Two Obj.} & \textbf{Counting} & \textbf{Colors} & \textbf{Position} & \textbf{Color Attri.} & \textbf{Overall} \\
\midrule
\multirow{8}{*}{\rotatebox{90}{Gen. Only}}
 & PixArt-$\alpha$~\cite{pixartalpha}     & 0.98 & 0.50 & 0.44 & 0.80 & 0.08 & 0.07 & 0.48 & 71.11 & - \\
 & SDv2.1~\cite{sd}               & 0.98 & 0.51 & 0.44 & 0.85 & 0.07 & 0.17 & 0.50 & 68.09 & -\\
 & Emu3-Gen~\cite{emu3}             & 0.98 & 0.71 & 0.34 & 0.81 & 0.17 & 0.21 & 0.54 & 80.60 & 46.02 \\
 & SDXL~\cite{sdxl}                 & 0.98 & 0.74 & 0.39 & 0.85 & 0.15 & 0.23 & 0.55 & 74.65 & 39.75\\
 & DALL-E 3~\cite{dalle3}           & 0.96 & 0.87 & 0.47 & 0.83 & 0.43 & 0.45 & 0.67 & 83.50 & 69.18\\
 & SD3-Medium~\cite{sd3}            & 0.99 & 0.94 & 0.72 & 0.89 & 0.33 & 0.60 & 0.74 & 84.08 & 60.71 \\
 & FLUX.1-dev~\cite{flux} & 0.98 & 0.93 & 0.75 & 0.93 & 0.68 & 0.65 & 0.82 & 84.00 & 61.30\\
\midrule
\multirow{9}{*}{\rotatebox{90}{Unified}}
 & TokenFlow-XL~\cite{tokenflow}    & 0.95 & 0.60 & 0.41 & 0.81 & 0.16 & 0.24 & 0.55 & 73.38 & -\\
 & Janus~\cite{janus}               & 0.97 & 0.68 & 0.30 & 0.84 & 0.46 & 0.42 & 0.61 & 79.68 & 51.23\\
 & Show-o~\cite{showo}              & 0.98 & 0.80 & 0.66 & 0.84 & 0.31 & 0.50 & 0.68 & 67.27 & -\\
 & Show-o2~\cite{showo2}              & 1.00 & 0.87 & 0.58 & 0.92 & 0.52 & 0.62 & 0.76 & 86.14 & 61.90\\
 & MetaQuery-XL~\cite{metaqueries} & - & - & - & - & - & - & 0.80 & 82.05 & - \\
 & Janus-Pro~\cite{januspro}     & 0.99 & 0.89 & 0.59 & 0.90 & 0.79 & 0.66 & 0.80 & 84.19 & 61.61\\
 & BLIP3-o~\cite{blip3o} & - & - & - & - & - & - & 0.84 & 81.60 & 59.87\\
 & BAGEL~\cite{bagel}                   & 0.99 & 0.94 & 0.81 & 0.88 & 0.64 & 0.63 & {0.82} & 84.03 & 61.53\\
 & BAGEL$^{\dagger}$~\cite{bagel} & 0.99 & 0.93 & 0.78 & 0.89 & 0.50 & 0.59 & 0.78 & 84.02 & 60.51\\
 \midrule
\multirow{4}{*}{\rotatebox{90}{Refine}}
 & BAGEL-RvE (UiG~\cite{uig})                  & 0.99 & 0.93 & 0.81 & 0.89 & 0.54 & 0.67 & {0.80} & 85.11 & 64.91\\
 & BAGEL-RvE (Uni-CoT~\cite{unicot})                  & 0.99 & 0.96 & 0.84 & 0.92 & 0.57 & 0.71 & {0.83} & 83.17 & 69.86\\
 & BAGEL-RvE (IRG~\cite{irg})                  & 0.98 & 0.94 & 0.83 & 0.86 & 0.74 & 0.73 & {0.85} & - & -\\
 \rowcolor{blue!10}
 & \textbf{BAGEL-RvR (Ours)}                  & \textbf{{1.00}} & \textbf{{0.96}} & \textbf{{0.91}} & \textbf{{0.93}} & \textbf{{0.86}} & \textbf{{0.80}} & \textbf{{0.91}} & \textbf{87.21} & \textbf{77.41}\\
\bottomrule
\end{tabular}}\vspace{-4mm}
\end{table}

\subsection{Multi-round generation}

As an iterative image refinement pipeline, RvR is expected to continuously improve results over multiple rounds. Here we investigate two key questions:

\begin{itemize}
    \item For \textbf{misaligned semantics} that remain after the first round, can additional RvR iterations further correct them?
    \item For \textbf{aligned semantics} already corrected in the first round, will additional iterations damage the correct results?
\end{itemize}

\begin{figure}[t]
    \centering
    \includegraphics[width=\linewidth]{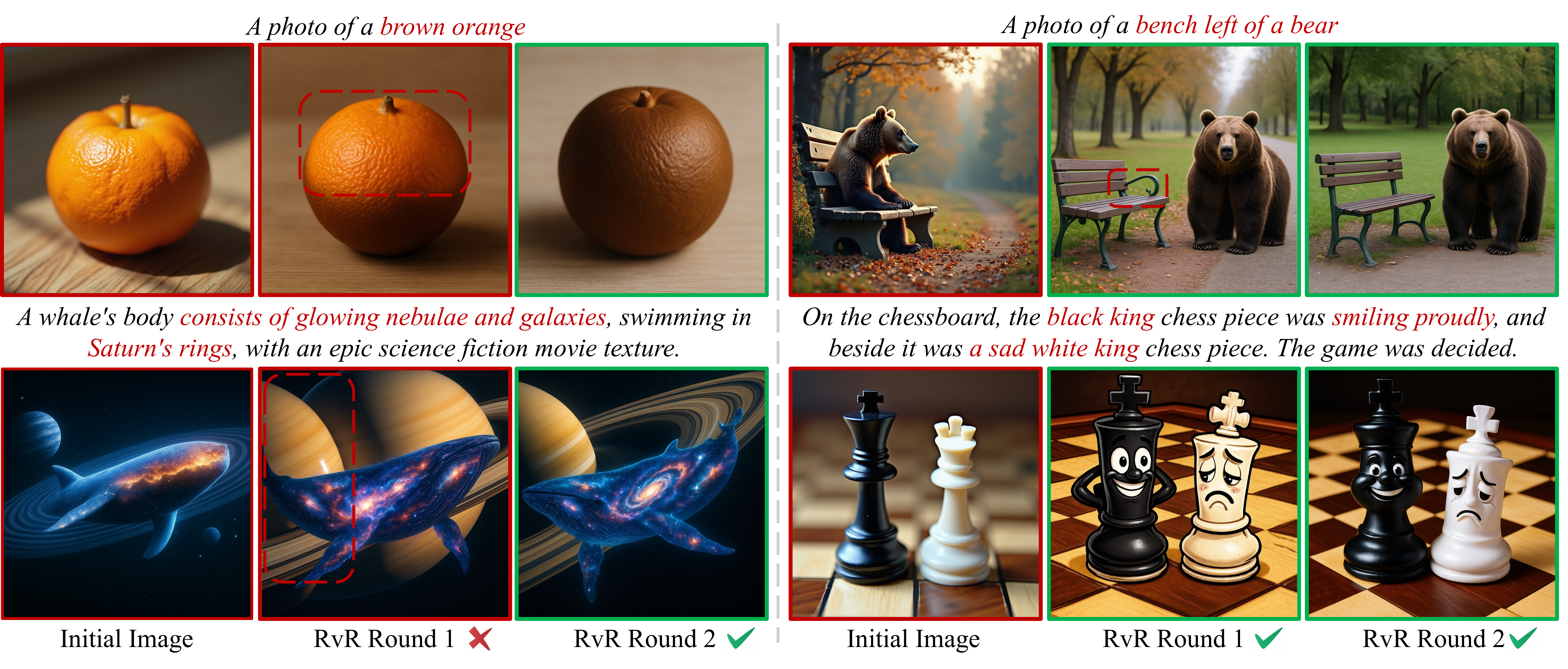}  \vspace{-6mm}
    \caption{\textbf{Multi-round generation.} RvR supports iterative refinement across multiple rounds. 
(a) Additional rounds can further correct misaligned semantics that remain unresolved after the first round. 
(b) When the semantics are already correctly aligned after the first round, another round preserves the correct results. 
    }\vspace{-4mm}
    \label{fig:f2}
\end{figure}

In~\cref{fig:f2}, we visualize several representative cases to answer these questions. The left three columns show examples where the visual results are still not well aligned with the desired prompts after the first round of RvR. By performing another round of refinement, the misaligned semantics are further corrected, \eg, the non-brown upper half of the orange and the unexpected duplicated Saturn appearing behind the whale. 
The right three columns present cases where we intentionally perform another round of RvR even though the first-round results are already aligned with the prompts. The second-round results show that the aligned semantics are well preserved. Meanwhile, the additional refinement can further improve minor visual details, \eg, the bench with only one armrest is refined into a more natural bench without armrests in the second round.

\subsection{Robustness to Initial Image Semantics}

As a robust image refinement pipeline, RvR is expected to reuse compatible semantics to ease regeneration while discarding conflicting ones to avoid unnatural compositions. To evaluate this robustness, we construct a \textbf{special experimental setting} where the initial image has a clear semantic gap from the desired prompt. We investigate two key questions:

\begin{itemize}
    \item For \textbf{prompt-compatible semantics} in the initial image, will RvR reuse them during regeneration?
    \item For \textbf{prompt-conflicting semantics}, will RvR discard them instead of forcing their preservation?
\end{itemize}

\begin{figure}[t]
    \centering
    \includegraphics[width=0.9\linewidth]{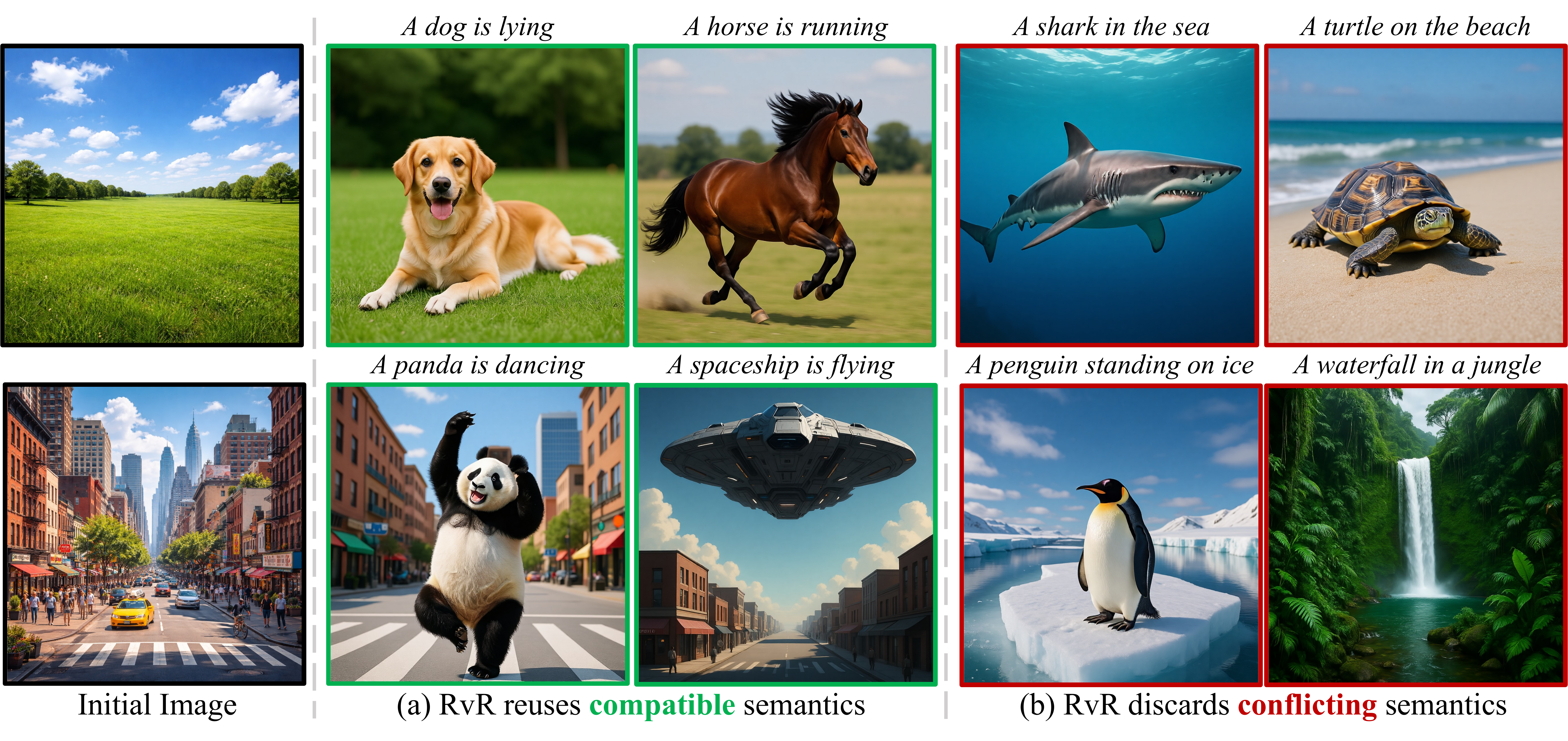}  \vspace{-3mm}
    \caption{\textbf{Robustness to initial image semantics.} 
(a) RvR reuses prompt-compatible semantics from the initial image to facilitate regeneration. 
(b) When the initial semantics conflict with the prompt, RvR discards them and regenerates new aligned content.}\vspace{-5mm}
    \label{fig:f1}
\end{figure}

These questions help reveal whether RvR truly leverages the semantics in the initial image during regeneration, rather than behaving like a stronger T2I pipeline that ignores the initial result.
In~\cref{fig:f1}, the first column shows initial images that are semantically far from the desired prompts. The second and third columns illustrate cases where the initial image contains semantics compatible with the prompts. For example, it is natural for a dog to lie on grass and reasonable for a spaceship to appear above a city. Although elements such as grass, trees, buildings, and streets are not explicitly mentioned in the prompts, they are preserved in the refined results. This suggests that RvR refers to the initial image and reuses compatible semantics during regeneration.

In contrast, the fourth and fifth columns present cases where the initial semantics strongly conflict with the prompt. For instance, ``a shark in the sea'' is incompatible with grass and trees, and ``a waterfall in a jungle'' is unlikely to appear in a city scene. In these cases, RvR discards the conflicting semantics and generates new images aligned with the prompts.
This behavior reflects the enlarged modification space of RvR: when compatible semantics exist, they are reused to simplify regeneration; when strong conflicts arise, the model discards them and regenerates a new aligned image, indicating strong pipeline robustness.

\subsection{Ablation Studies}\label{sec:abl}

In~\cref{tab:abl}, we analyze the impact of different training strategies and design choices on DPGBench. The key findings are summarized below.

\smallskip
\noindent\textbf{RvR primarily benefits from refinement training.}
During training, RvR incorporates T2I data to preserve the basic T2I capability of BAGEL (\cref{sec:setup}). We therefore evaluate the T2I performance after RvR training (BAGEL-RvR-T2I). The results show performance comparable to the original BAGEL model (84.08 vs. 84.02), indicating that the T2I capability is well preserved. Meanwhile, the overall performance gain mainly comes from the refinement training process rather than improvements in T2I generation.

\smallskip
\noindent\textbf{RvR outperforms SFT with the same training data scale.}
By discarding misaligned images, we convert refinement triplets $\langle$misaligned image, aligned image, prompt$\rangle$ into T2I pairs $\langle$aligned image, prompt$\rangle$ and conduct SFT on BAGEL with the same data scale (BAGEL-SFT). This setting yields only a minor improvement over BAGEL (84.62 vs. 84.02). The result suggests that the performance gain of RvR mainly comes from the regeneration-based refinement mechanism rather than simply using higher-quality finetuning data. 

\smallskip
\noindent\textbf{Editing data degrades the performance.}
We further evaluate the effect of incorporating additional editing data into RvR training (BAGEL-RvR + Editing Data). To align with the RvR pipeline, we replace the editing instructions with target prompts. However, this leads to a performance drop (85.70 vs. 87.21). A possible reason is that the strong pixel-level consistency between source and target images in editing data encourages the model to pursue pixel-level preservation, which restricts the modification space for effective semantic correction.

\smallskip
\noindent\textbf{VAE degrades the performance.}
We also test incorporating a VAE to encode pixel-level features of the input image in RvR (BAGEL-RvR + VAE). However, such features are largely irrelevant to the goal of semantic correction in RvR. As a result, introducing these features slightly harms performance (86.41 vs. 87.21).

\begin{table}[t]
    \centering
    \caption{\textbf{Ablation studies on RvR training strategies and design choices.} 
We compare refinement training with alternative strategies (T2I and SFT) and examine the effect of incorporating editing data and VAE features on DPGBench.} \vspace{-2.5mm}
    \label{tab:abl}
    \setlength{\tabcolsep}{4pt}
    \resizebox{0.9\linewidth}{!}{
    \begin{tabular}{lcccccc}
        \toprule
      \multirow{2}{*}{\textbf{Setting}} & \multicolumn{6}{c}{\textbf{DPGBench↑}}\\
      & Global & Entity & Attribute & Relation & Others & Overall \\
        \midrule
        BAGEL  & 91.55 & 89.95 & 89.87 & 89.22 & 88.90 & 84.02\\
        \midrule 
        \textcolor{gray}{\textit{\textbf{RvR vs. T2I/SFT}}}\\
        BAGEL-RvR-T2I & 90.88 & 88.81 & 91.07 & 88.59 & 87.51 & 84.08\\
        BAGEL-SFT  & 89.19 & 90.48 & 90.09 & 90.82 & 91.04 & 84.62\\
         \midrule 
         \textcolor{gray}{\textit{\textbf{RvR with editing components}}}\\
         BAGEL-RvR + Editing Data & 91.53 & 90.88 & 91.27 & 90.88 & 89.81 & 85.70 \\ 
         BAGEL-RvR + VAE & 91.21 & 91.61 & 91.81 & 91.99 & 89.56 & 86.41 \\
         \midrule 
         \rowcolor{blue!10}BAGEL-RvR (Ours) & \textbf{{91.91}} & \textbf{{91.74}} & \textbf{{91.84}} & \textbf{{92.66}} & \textbf{{92.09}} & \textbf{87.21} \\ 
        \bottomrule
    \end{tabular}}\vspace{-3.5mm}
\end{table}

\section{Conclusion}
\label{sec:con}

In this paper, we revisited image refinement in unified multimodal models from a regeneration perspective. We showed that editing-based refinement constrains the modification space through editing instructions and pixel-level consistency requirements, limiting prompt--image alignment performance.
To address this issue, we proposed \emph{Refinement via Regeneration} (RvR), which reformulates refinement as another round of generation conditioned on the target prompt and semantic tokens of the input image. By discarding editing instructions and unnecessary consistency constraints, RvR enables more flexible semantic correction.
We further introduced a training paradigm based on independently generated images with different prompt-alignment levels, encouraging semantic correction rather than appearance preservation. Experiments across multiple benchmarks demonstrate that RvR consistently improves text-to-image generation performance across various benchmarks. These results suggest that regeneration, rather than editing, provides a more effective foundation for refinement.

\bibliographystyle{splncs04}
\bibliography{main}

\clearpage
\setcounter{page}{1}

\appendix

\section*{Supplementary Materials}

\section{Attention Mask}

We adopt the standard omni-attention mechanism~\cite{showo} in UMMs to support RvR training for image refinement. Causal attention is applied to text tokens from the system prompt and T2I prompt, whereas full attention is applied to ViT tokens of the misaligned image and noisy VAE tokens of the aligned image.

\begin{figure}[h]
    \centering\vspace{-4mm}
    \includegraphics[width=0.5\linewidth]{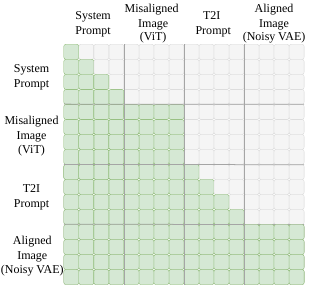} 
    \caption{\textbf{Attention mask for RvR.} 
We follow standard UMM training to apply causal attention for text tokens (prompts) and full attention for image tokens (ViT and VAE). }\vspace{-8mm}
    \label{fig:a1}
\end{figure}

\section{Refinement Data Examples}

\begin{figure}[h]
    \centering\vspace{-4mm}
    \includegraphics[width=\linewidth]{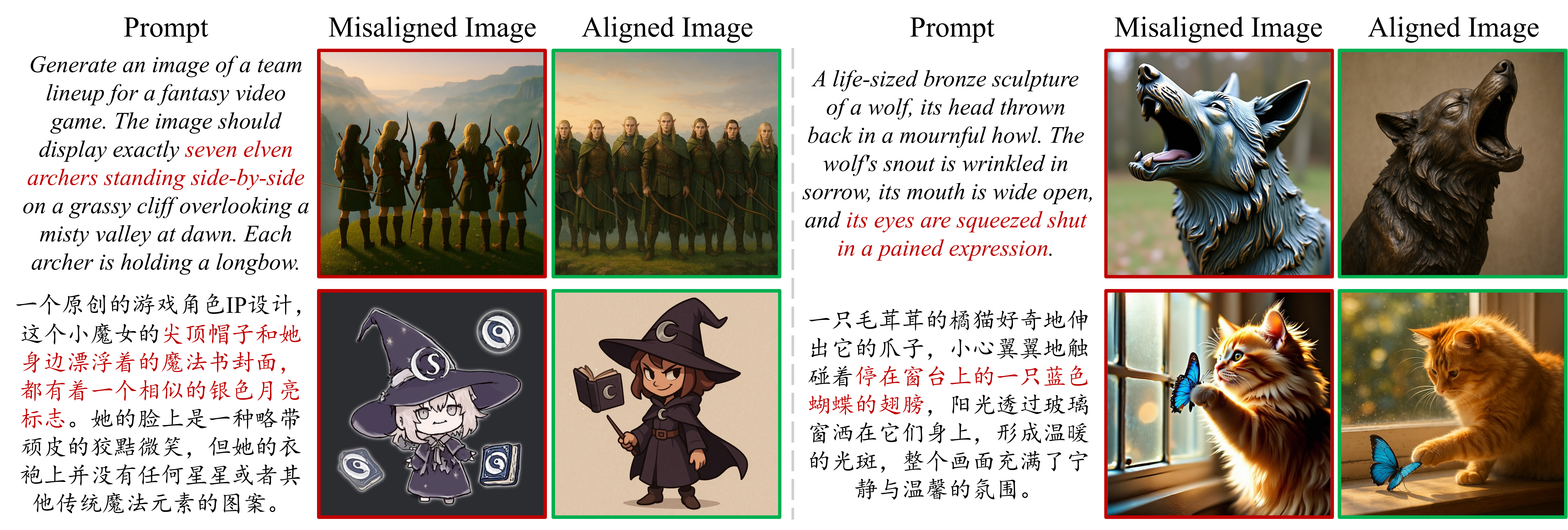} 
    \caption{\textbf{Sampled training triplets for RvR.} The prompts are constructed with a 1:1 ratio of English and Chinese, allowing RvR to support bilingual refinement.}\vspace{-6mm}
    \label{fig:a2}
\end{figure}

\cref{fig:a2} visualizes several refinement data examples used for RvR training. 
Unlike image editing, the misaligned and aligned images are independently generated by the UMM's T2I process. 
This independence removes unnecessary pixel-level constraints and encourages RvR to focus on semantic correction.

\section{Additional Qualitative Results}

\begin{figure}[h]
    \centering\vspace{-2mm}
    \includegraphics[width=0.92\linewidth]{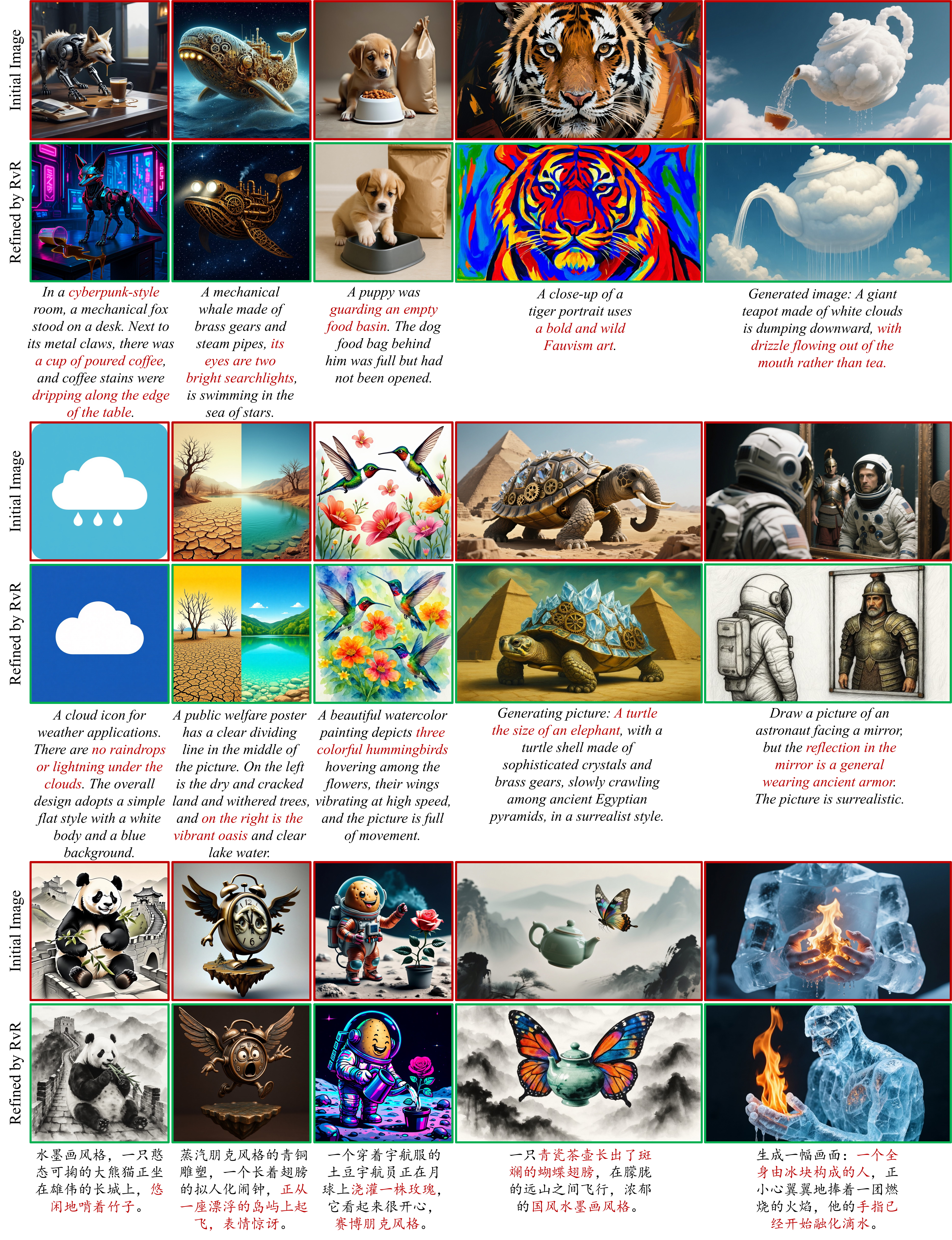} 
    \caption{\textbf{Additional qualitative results.}}\vspace{-6mm}
    \label{fig:a3}
\end{figure}

\cref{fig:a3} presents additional qualitative results further demonstrating the refinement performance of RvR, including examples with Chinese prompts to showcase the bilingual refinement capability.

\clearpage
\end{document}